\documentclass[12pt, a4paper]{article}

\usepackage{subcaption}     
\usepackage{authblk}

\usepackage{graphicx}
\usepackage{float}
\usepackage{array}      
\usepackage{booktabs}    
\usepackage{amsmath}     

\usepackage{geometry}
\usepackage{cite}

\usepackage[colorlinks=true, linkcolor=blue, citecolor=blue]{hyperref}

\title{TianJi：An autonomous AI meteorologist for discovering physical mechanisms in atmospheric science}

\author[1]{Kaikai Zhang}
\author[1]{Xiang Wang}
\author[1]{Haoluo Zhao}
\author[1]{Nan Chen}
\author[1]{Mengyang Yu}
\author[1,2]{Jing-Jia Luo}
\author[3]{Tao Song}
\author[1,2,*]{Fan Meng}

\affil[1]{School of Artificial Intelligence, Nanjing University of Information Science and Technology, Nanjing, China}
\affil[2]{State Key Laboratory of Climate System Prediction and Risk Management (CPRM), Nanjing University of Information Science and Technology, Nanjing, 210044, China}
\affil[3]{College of Computer Science and Technology, China University of Petroleum, Qingdao, Shandong, China }
\affil[*]{Corresponding author: Fan Meng. E-mail: meng@nuist.edu.cn}

\date{\today}

\begin{document}
	
	\maketitle
	
\begin{abstract}
	Artificial intelligence (AI) has achieved breakthroughs comparable to traditional numerical models in data-driven weather forecasting, yet it remains essentially statistical fitting and struggles to uncover the physical causal mechanisms of the atmosphere. Physics-oriented mechanism research still heavily relies on domain knowledge and cumbersome engineering operations of human scientists, becoming a bottleneck restricting the efficiency of Earth system science exploration. Here, we propose TianJi—the first "AI meteorologist" system capable of autonomously driving complex numerical models to verify physical mechanisms. Powered by a large language model-driven multi-agent architecture, TianJi can autonomously conduct literature research and generate scientific hypotheses. We further decouple scientific research into cognitive planning and engineering execution: the meta-planner interprets hypotheses and devises experimental roadmaps, while a cohort of specialized worker agents collaboratively complete data preparation, model configuration, and multi-dimensional result analysis. In two classic atmospheric dynamic scenarios (squall-line cold pools and typhoon track deflections), TianJi accomplishes expert-level end-to-end experimental operations with zero human intervention, compressing the research cycle to a few hours. It also delivers detailed result analyses and autonomously judges and explains the validity of the hypotheses from outputs. TianJi reveals that the role of AI in Earth system science is transitioning from a "black-box predictor" to an "interpretable scientific collaborator", offering a new paradigm for high-throughput exploration of scientific mechanisms.
	
	\vspace{1em}
\end{abstract}
	
	\section{Introduction}
	Artificial intelligence has achieved breakthroughs comparable to traditional numerical models in data-driven weather forecasting. Models such as PanguWeather and GraphCast have attained forecast accuracy that matches or even surpasses traditional numerical weather prediction for medium-range and short-range global weather forecasts through end-to-end statistical mapping, representing the tremendous progress of data-driven methods in atmospheric state prediction tasks\cite{article1, article2,chen2023fuxi,xiao2024coupling,luo2026ai}.
	
	However, predictive capability is not equivalent to scientific understanding. The essence of these models is high-dimensional statistical fitting of atmospheric state time series, which completely bypasses the explicit modeling of atmospheric physical mechanisms. They can predict weather states but cannot answer the profound "why" questions in atmospheric science\cite{vu2025black,mengaldo2024explain}. Answering such questions must rely on controlled experiments centered on physical numerical models, making it urgent to integrate AI with physical simulators such as WRF. Existing AI agent systems for climate science primarily focus on automated data analysis and statistical diagnostics\cite{guo2025self}. They have not yet achieved the autonomous driving of mesoscale numerical models like WRF to verify atmospheric physical mechanisms. The TianJi system proposed in this paper fills this critical research gap.
	\par
	Physical simulators have an extremely high barrier to use. Constructing a single agent requires mastering cross-modal, multi-step capabilities including atmospheric dynamics, HPC scheduling, Fortran configuration, and NetCDF data analysis, which inevitably leads to hallucinations or breakdowns when handling ultra-long context tasks. Meanwhile, physics-oriented mechanism research heavily depends on the domain knowledge and cumbersome engineering operations of human scientists, becoming a bottleneck restricting the efficiency of Earth system science exploration\cite{schneider2023harnessing}.
	\par
	Scientific research is essentially a social process. Therefore, instead of adopting a more complex single model, we built a "virtual research group" multi-agent architecture\cite{zhou2025autonomous,lyu2026evoscientist}: we decouple scientific research into cognitive planning and engineering execution, where the meta-planner takes charge of directional guidance, and specialized worker agent groups handle execution details. Through role separation and cross-validation, we bridge the gap between natural language and complex Fortran physical systems.
	\par
	Based on the above research ideas, we propose TianJi\footnote{We name our AI meteorologist TianJi. 
		In Chinese philosophy, ``Tian'' refers to the sky or heaven, and ``Ji'' denotes mechanisms or profound secrets. Historically, 
		forecasting the weather was considered attempting to decipher ``TianJi''---the unrevealable secrets of nature. 
		Here, our model systematically unravels these atmospheric mechanisms through artificial intelligence.}—the first AI meteorologist system capable of autonomously driving complex numerical models to verify physical mechanisms. Driven by large models and centered on WRF physical simulation, the system consists of two core modules: hypothesis generation and hypothesis verification, which can fully support the entire research process from scientific hypothesis to mechanism validation.
	\par
	Among them, the hypothesis generation module adopts a multi-agent design centered on a debate mechanism. By simulating the peer review and opinion confrontation mechanisms in academic discussions\cite{shahhosseini2025large,lin2025creativity}, combined with paper retrieval and multi-round iterative optimization, it automatically proposes research hypotheses with innovation and scientificity.
	The hypothesis verification module decouples the research process relying on a multi-agent architecture, effectively solving the core pain points that existing meteorological language models cannot call external physical simulators or process NetCDF files, and ultimately realizes the core transformation of AI in atmospheric science from a "black-box predictor" to an "interpretable scientific collaborator".\footnote{The test results and subsequent open-source code are available at \url{https://github.com/zwww-www/output}.}
	
	\section{System Architecture}
\begin{figure}[H]
		\centering
		\includegraphics[width=1\textwidth]{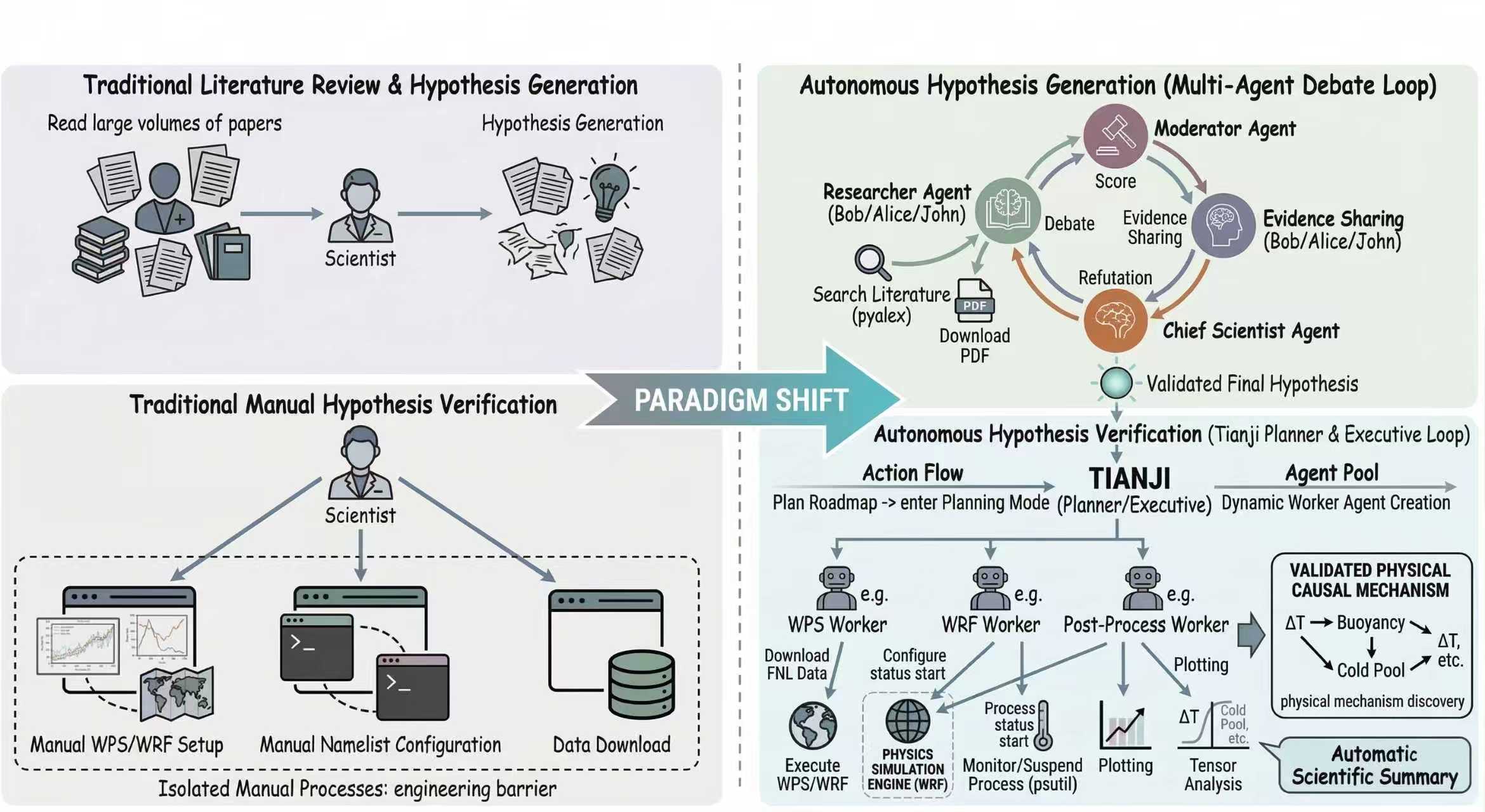} 
		\caption{System Architecture Diagram}
		\label{img1}
	\end{figure}

	The TianJi system is composed of two parts: hypothesis generation and hypothesis verification, both of which are built based on the AgentScope framework\cite{gao2025agentscope}. These two components jointly constitute the key functional modules for scientific research exploration, supporting the entire research process from literature investigation to experimental verification. The overall workflow is illustrated in Figure\ref{img1}.
	
	\subsection{Hypothesis Generation: Iterative Optimization System Based on Academic Seminar-Style Debate}
	In recent years, large language models (LLMs) have demonstrated great potential in scientific exploration, yet direct generation of scientific hypotheses using LLMs still faces numerous challenges. Additionally, LLMs are prone to "academic hallucinations" when generating content directly, leading to hypotheses that seem reasonable on the surface but lack factual accuracy and scientific grounding, which severely compromises their reliability in research\cite{farquhar2024detecting}.
	\par
	To address the limitations of direct generation methods, some studies have adopted single-agent architectures with self-evaluation mechanisms to improve the novelty and relevance of hypotheses\cite{shinn2023reflexion}. However, single-agent approaches cannot reproduce the collaborative nature of real-world scientific research, lacking the collision and divergence of multidisciplinary perspectives, which restricts the depth and breadth of solving complex scientific problems. In contrast, multi-agent collaboration methods can simulate the peer review and team discussion processes in the scientific community\cite{lu2024ai}, and significantly boost the overall novelty, feasibility, and theoretical depth of generated hypotheses through collaboration and debate among agents with diverse viewpoints\cite{du2024improving}.
	\par
	Current LLM-driven scientific hypothesis generation relies on four core methods: iterative refinement, retrieval augmentation, multi-agent collaboration, and multimodal fusion. Integrating mainstream existing hypothesis generation techniques, we designed and built a hypothesis generation system centered on an academic seminar-style debate mechanism, coupled with a closed loop of continuous iteration and dynamic tuning. This system adopts a multi-agent collaborative architecture composed of researcher agents, a host agent, and a chief scientist agent. All three types of agents share full-domain information and complete hypothesis generation via structured debate and iterative optimization. The core operating logic and configurable features are as follows:
	\begin{enumerate}
		\item Researcher agents act as the core participants in the debate. Their quantity can be flexibly adjusted manually, and each researcher agent can be assigned differentiated domain expertise (e.g., thermodynamics, dynamics) to mine hypotheses from multidisciplinary perspectives for the same research topic. The total number of debate rounds and the speaking order of researcher agents can also be freely arranged and customized.
		\item The system integrates a closed loop of multi-round debate–evaluation–optimization. After each debate round, the host agent quantitatively scores the hypotheses proposed by researcher agents across four dimensions: scientificity, rationality, novelty, and effectiveness\cite{chan2023chateval}. Researcher agents optimize their hypotheses in targeted ways in subsequent rounds based on these scores.
		\item The system features a rebuttal mechanism that allows researcher agents to refute each other’s hypotheses by pointing out flaws and irrationality. Rebutted agents speak first and selectively accept and revise their hypotheses based on the rebuttals, enhancing the focus and depth of the debate.
		\item After all debate and iteration processes are completed, the chief scientist agent comprehensively reviews and screens all optimized hypotheses to determine the final hypothesis.
	\end{enumerate}
	\par
	Before the debate starts, users can freely specify the scientific topic to be discussed. In the first round, researcher agents do not debate or refute, but propose their own hypotheses. Before proposing hypotheses, the system retrieves the titles and abstracts of relevant papers via a paper retrieval tool to assist hypothesis generation\cite{gao2023retrieval}, making hypotheses more scientific and theoretically grounded. In the subsequent iterative rounds, researcher agents officially launch cross-rebuttal and structured debate, focusing on pointing out defects and limitations in each other’s schemes. Throughout the system lifecycle, full-domain information remains absolutely transparent, ensuring every agent can obtain and coordinate all historical interaction information in real time. Users can independently add fact-enhancement tools during use to provide more external knowledge.
	
	\subsection{Hypothesis Verification: WRF-Centered Multi-Agent System}
	In recent years, large language model (LLM)-driven multi-agent systems have shown great potential in automating complex physical simulations and have been successfully applied to multiple frontier disciplines. For instance, in the field of computational fluid dynamics (CFD), systems such as CFDagent have achieved end-to-end zero-shot automated simulation from geometry generation to flow field solution and result visualization\cite{xu2025cfdagent}. Foam-Agent 2.0 further demonstrates how to realize a closed-loop full-chain simulation by encapsulating independent agents for meshing, running, and visualization\cite{yue2025foam}. Multi-agent reinforcement learning has also been successfully applied to active flow control in turbulent environments\cite{vasanth2025multi}. In microscopic molecular dynamics (MD) calculations, frameworks such as DynaMate and other atomic-level simulation systems have achieved collaboration from parameter configuration, cluster (HPC) scheduling to visual feedback-based anomaly self-correction\cite{mendible2025dynamate}. In addition, the PhysAgent system has even proven that multi-agent architectures can autonomously schedule first-principles calculation tools and derive physical laws\cite{han2025physagent}.
	\par
	However, although multi-agent architectures have excelled in rigorous scientific computing such as fluid and molecular simulations, their application in macroscopic meteorological simulation—especially atmospheric dynamic simulation represented by the complex Weather Research and Forecasting (WRF) model—remains a gap. To fill this research gap, this paper proposes for the first time a deep integration of multi-agent collaboration mechanisms with WRF simulation to construct an automated meteorological scientific hypothesis verification system.
	\par
	The hypothesis verification module is a WRF-centered multi-agent execution engine. To adapt to verification requirements of varying complexity, the system supports two execution modes, which can be automatically selected based on the complexity of the input task or manually specified by the user:
	\begin{enumerate}
		\item Simple Mode: For low-complexity tasks such as analyzing single results and plotting basic meteorological fields. This mode skips complex task planning and scheduling steps, allowing agents to directly invoke underlying tools in single or multiple steps for rapid response.
		\item Complex Mode: For complex verification tasks requiring multi-step reasoning and multi-dimensional analysis. Drawing on the hierarchical scheduling design of advanced multi-agent systems, the master agent is responsible for logical decomposition, coordination, and dynamic scheduling of upper-level complex tasks\cite{boiko2023emergent}. The master agent dynamically creates dedicated sub-agents in real time according to the task pipeline, which complete corresponding sub-tasks.
	\end{enumerate}
	\par
	To support agents in completing end-to-end meteorological verification tasks, following the paradigm of modular tool encapsulation in advanced computing frameworks\cite{wang2023scientific}, the system customizes four categories of underlying tool buses for the meteorological domain:
	\begin{enumerate}
		\item Physical Simulation Module: As the core driver of the system, it is exclusively responsible for the underlying driving of the WRF model, the setting of the simulation computational domain, and the dynamic configuration of complex physical parameterization schemes.
		\item Spatiotemporal Tensor Calculation and Analysis Module: Specialized in processing high-dimensional gridded meteorological field data output by the WRF model, providing multi-dimensional tensor reading and aggregation, spatial geometric filtering, projection coordinate system conversion, automatic positioning and tracking of specific weather systems (e.g., cyclone centers), profile feature extraction, and various spatiotemporal statistical and analytical calculations.
		\item Visualization Module: Dedicated to rendering numerical calculation results and generating spatial distribution maps integrated with real geographic information (GIS) or intuitive multi-dimensional spatiotemporal statistical charts.
		\item Basic Tool Module: Encapsulates underlying system-level interaction capabilities, including read and write operations of basic configuration files, terminal command execution, and Python data processing code.
	\end{enumerate}
	\par
	This module is a WRF-centered multi-agent system. It can switch to Simple Mode to handle simple tasks and act as an assistant, or enter Complex Mode to function as a meteorologist. The four categories of underlying tool buses provide comprehensive technical support for the stable operation of both modes, ensuring efficient execution of meteorological hypothesis verification tasks of varying complexity. In Simple Mode, the system can quickly respond to the basic operational needs of researchers, such as rapidly retrieving WRF simulation results, plotting basic meteorological element distribution maps, and simply analyzing meteorological field characteristics, without requiring researchers to have complex WRF model operation experience or programming skills. This greatly lowers the entry barrier for meteorological simulation and hypothesis verification, serving as a capable assistant for researchers to conduct preliminary studies efficiently. In Complex Mode, through task decomposition by the master agent and collaborative work of sub-agents, the system can simulate the research thinking and analysis process of senior meteorologists, conduct end-to-end verification for complex meteorological hypotheses (e.g., causes of extreme precipitation, cyclone evolution mechanisms), and autonomously complete parameterization scheme configuration, high-dimensional data mining, multi-dimensional result analysis, and visualization presentation. It truly realizes an intelligent and automated closed loop for meteorological hypothesis verification, effectively compensating for the shortcomings of traditional WRF simulation that relies on manual operation, has low efficiency, and requires extremely high professional literacy of operators.
	
	\section{Experimental Validation}
	To verify the feasibility of the system, for the hypothesis generation module, this paper selects a topic for in-depth debate; for the hypothesis verification module, we validate scientific hypotheses in two typical complex atmospheric dynamic scenarios and four simple analysis tasks.
	\subsection{Hypothesis Generation Validation}
	For the experimental validation of hypothesis generation, the discussion topic selected for the experiment is as follows:
	\begin{quote}
		During the translation of a tropical cyclone (TC), complex interactions between the vortex and its environment can lead to sudden, anomalous track deflections (such as abrupt turns, stalling, or looping) that severely challenge forecasting models. Please propose a specific physically based mechanistic hypothesis to explain how the nonlinear interactions among large-scale environmental steering flows (e.g., subtropical high variations, mid-latitude troughs), internal vortex dynamics (e.g., diabatic heating asymmetries, beta effect), and air-sea coupling (e.g., ocean cold wake generation) trigger such anomalous typhoon track changes.
	\end{quote}
	Three researcher agents were configured in the experiment, with the built-in debate and rebuttal mechanism enabled and no additional adjustment to the speaking order. The experimental procedure is illustrated in Figure\ref{img2}:
	\begin{figure}[H]
		\centering
		\includegraphics[width=0.6\textwidth]{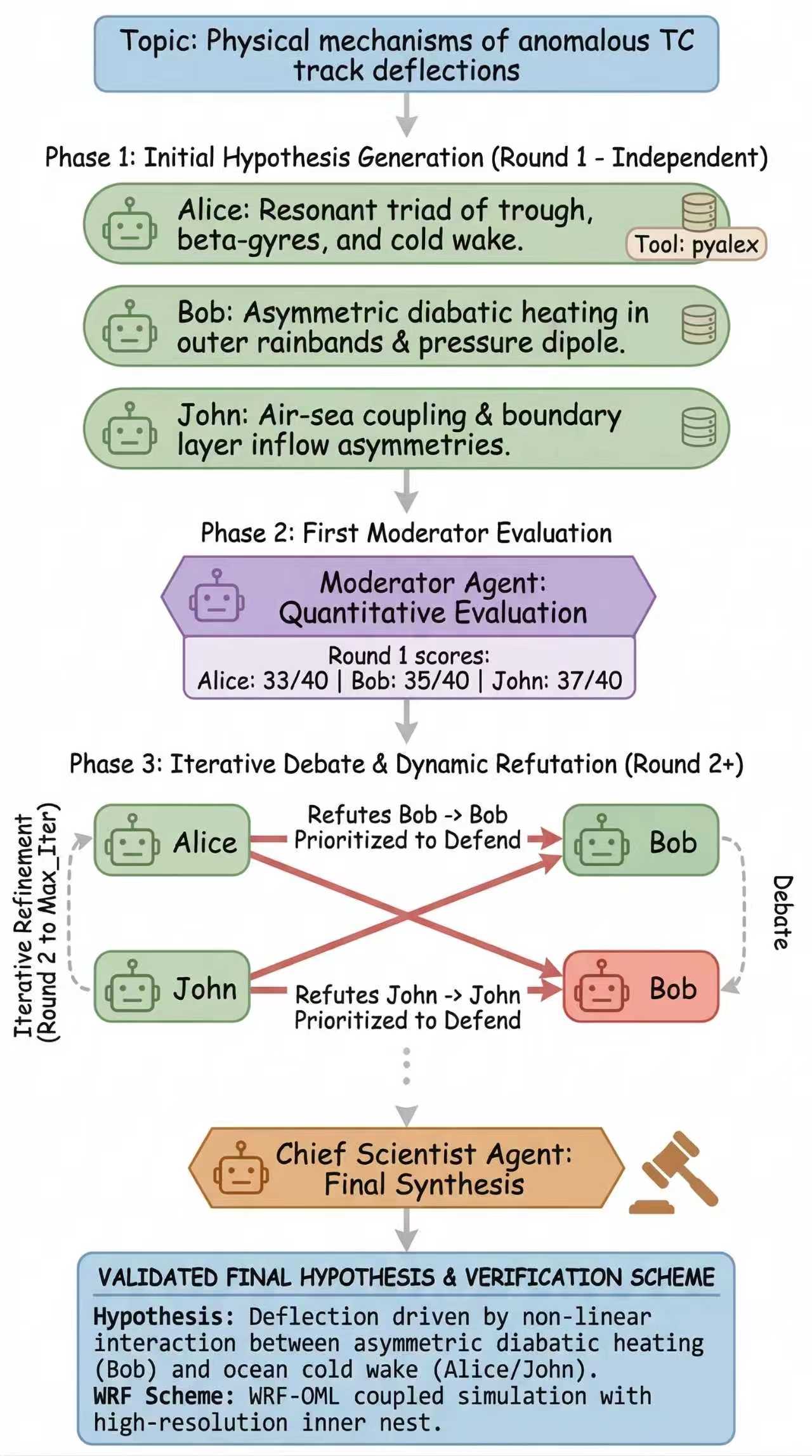} 
		\caption{Workflow Diagram of Hypothesis Generation}
		\label{img2} 
	\end{figure}
	After six rounds of iterative optimization, each agent proposed its own hypothesis, and the final scientific hypothesis was determined.
	\par
\begin{figure}[H]
		\centering
		\includegraphics[width=0.8\textwidth]{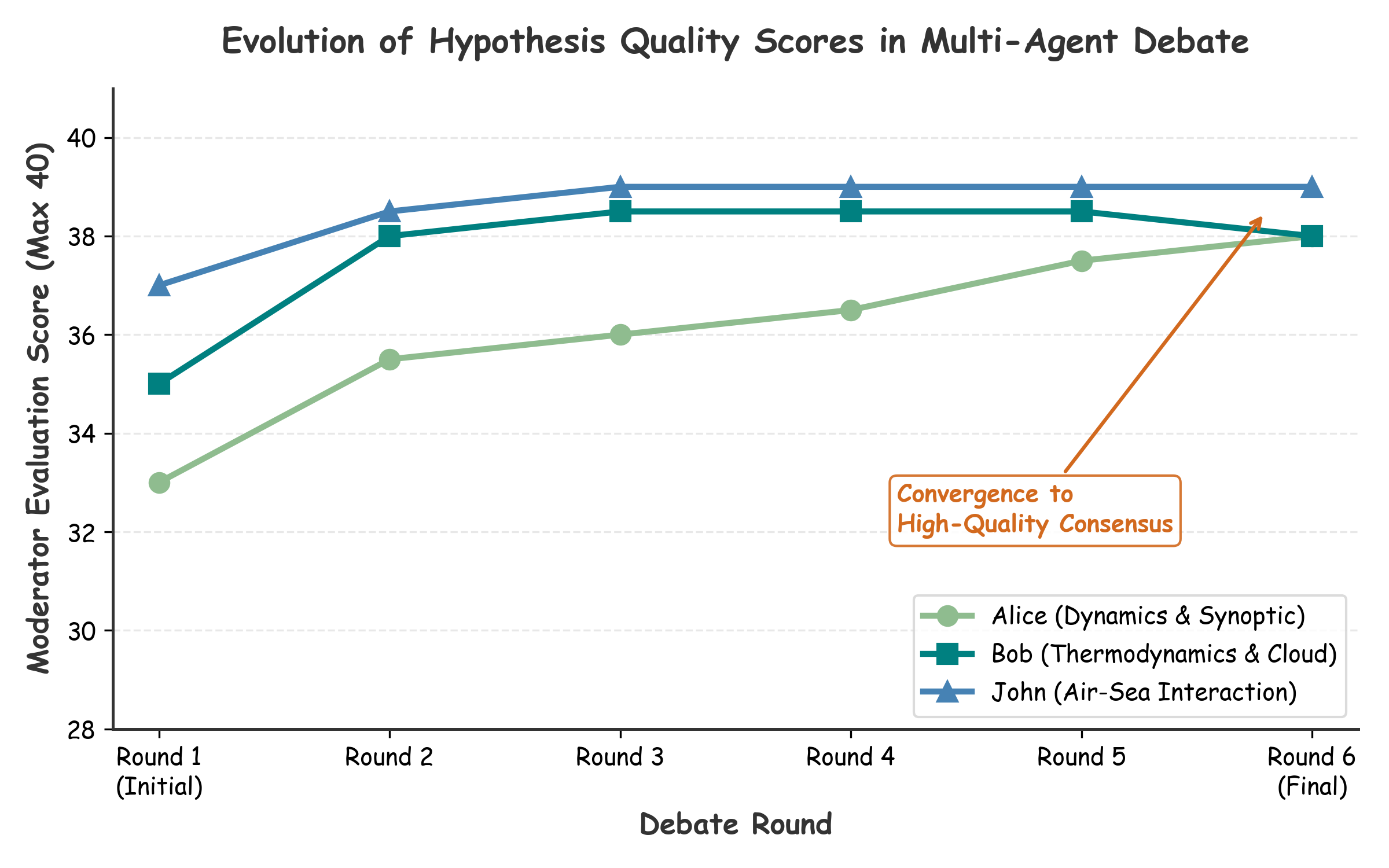} 
		\caption{Evolution Curve of Hypothesis Quality Scores During Multi-Agent Debate}
		\label{img3} 
	\end{figure}

	This experiment also evaluated the iterative optimization process, as shown in Figure\ref{img3}. This figure presents the convergence and quality improvement of the multi-agent debate mechanism. All researcher agents refined their hypotheses during the debate, and their scores showed a continuous upward trend. For instance, Alice’s score rose from a minimum of 33 to 38. Starting from the third round, the scores of the three researcher agents gradually approached the full score and stabilized, which strongly proves that the system did not fall into endless unstructured arguments, but successfully converged to a high-quality scientific consensus.
	
\subsection{Validation of the Hypothesis that Low Soil Moisture Enhances the Cold Pool Gust Front of Squall Lines}
\begin{figure}[htbp]        
	\centering              
	\begin{subfigure}[b]{0.45\textwidth}
		\centering       
		\includegraphics[width=\linewidth]{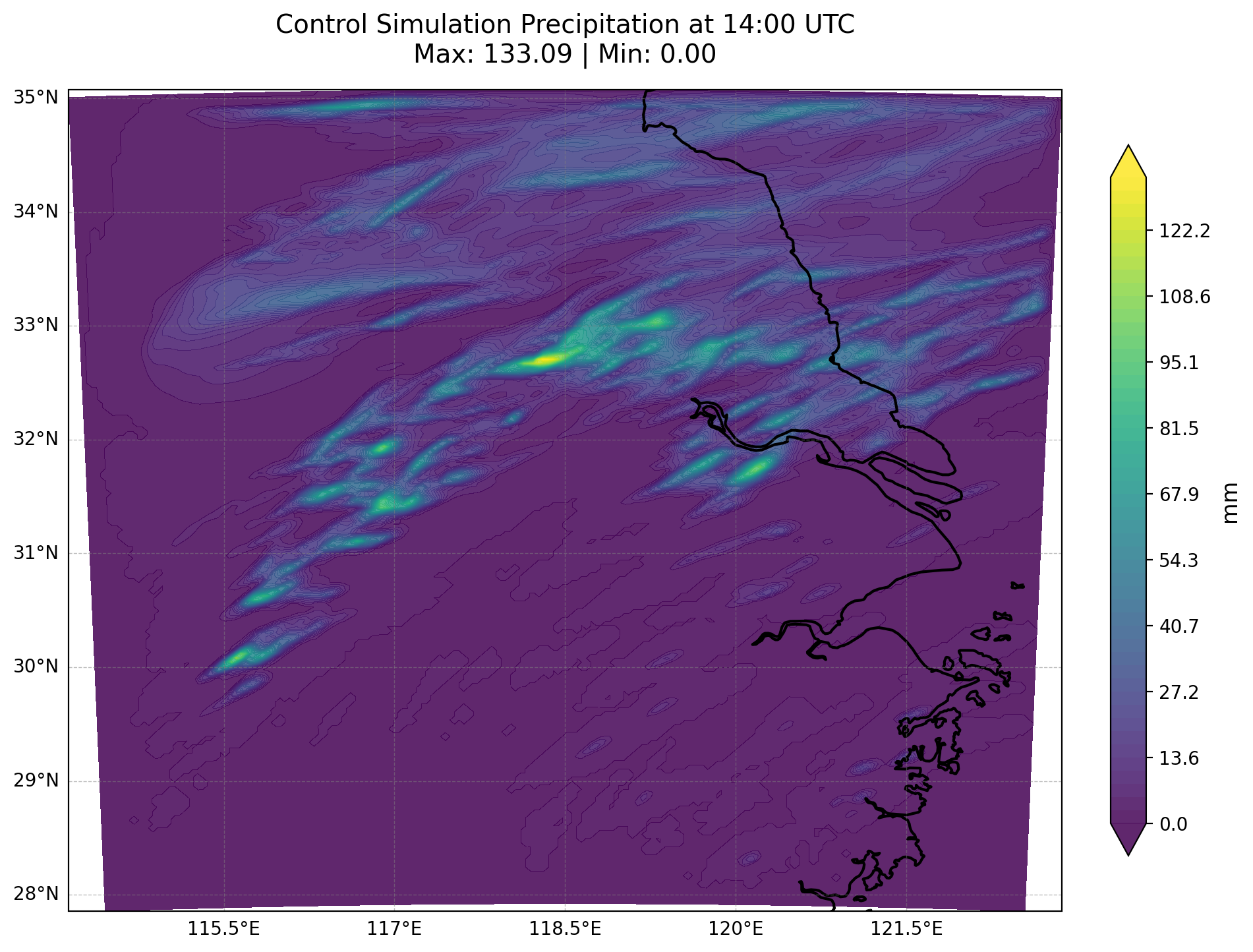}  
		\caption{Spatial distribution of squall line precipitation in the control group}        
		\label{img4}                        
	\end{subfigure}
	\hfill                 
	\begin{subfigure}[b]{0.45\textwidth}
		\centering
		\includegraphics[width=\linewidth]{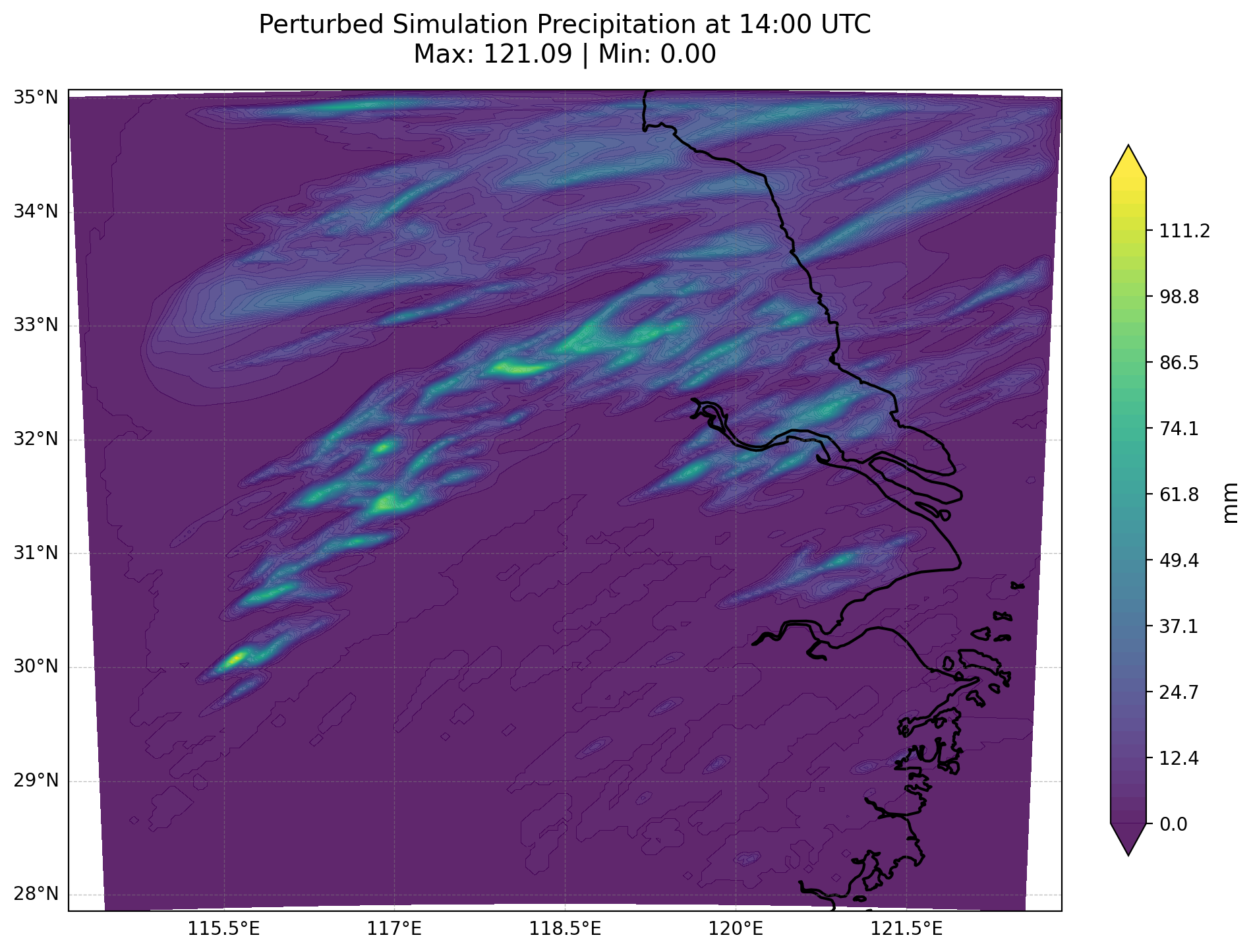}
		\caption{Spatial distribution of squall line precipitation under low soil moisture}
		\label{img5}
	\end{subfigure}
	\caption{Spatial distribution of precipitation}
	\label{img4_img5}
\end{figure}

In this experimental validation, the hypothesis tested is that abnormally dry soil (low soil moisture) ahead of the moving path of a squall line system significantly strengthens the intensity of the cold pool gust front, thereby accelerating the movement speed and prolonging the lifespan of the squall line system\cite{yang2026distinct}. After parsing the natural language research instruction, the TianJi system autonomously constructed the underlying physical mechanism chain and coordinated the design of the experimental scheme. Notably, the system independently selected the Noah-MP land surface model and the YSU boundary layer scheme—a configuration that enables fine characterization of the soil moisture-flux coupling process\cite{yu2025numerical}. The simulation results autonomously extracted by the TianJi system strongly validate this core hypothesis: compared with the control group, the cold pool characteristics of the soil moisture perturbation experimental group are significantly weakened, the temperature deficit within 50 km of the squall line core area is reduced by 20.4\% relative to the control group, and the squall line loses the coherent northeastward propagation feature observed in the control group due to structural fragmentation. As shown by the radar echo/total precipitation comparison in Figure\ref{img4_img5}, the squall line loses the coherent northeastward propagation feature of the control group due to structural fragmentation.

	\par
	Most importantly, the analysis report of the TianJi system goes far beyond mere hypothesis validation. This AI meteorologist autonomously identified emerging secondary signals that were not emphasized in the researchers' initial analysis by integrating multi-dimensional simulation outputs including 2-m air temperature, 10-m wind field, and accumulated precipitation: the spatial coherence of the squall line in the soil moisture perturbation experimental group was significantly attenuated, and the main convective activity shifted southward by approximately 280 km. The TianJi system also reasonably speculated in the output report that this phenomenon stems from the reduced evaporative cooling efficiency caused by changes in soil moisture, which further leads to an imbalance in the dynamic feedback between the cold pool and the environmental airflow—a detail that was not fully considered in the original research hypothesis focusing on thermodynamic processes\cite{fu2025impact}. This result confirms that the TianJi system has the ability to mine hidden meteorological signals, and the system has marked "the dynamic mechanism of squall line spatial displacement under soil moisture perturbation" as the research objective for subsequent independent experiments.
	\par
\begin{figure}[H]
		\centering
		\includegraphics[width=0.8\textwidth]{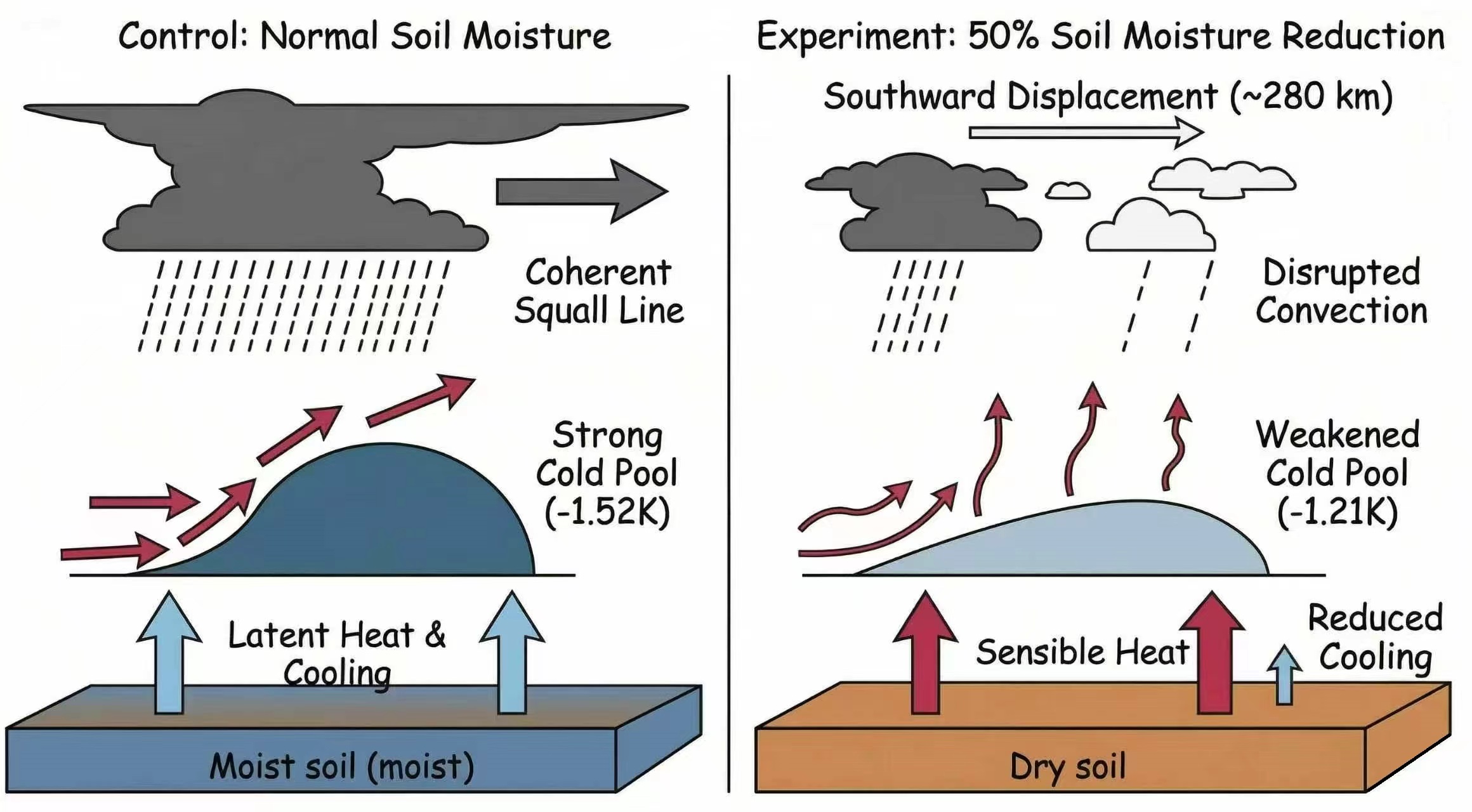} 
		\caption{Mechanism of Low Soil Moisture Impact on Squall Lines}
		\label{img6}
	\end{figure}

	The above simulation results reveal the multi-dimensional effects of soil moisture perturbation on squall line systems. Combined with Figure\ref{img6}, the underlying physical regulatory mechanisms can be further decomposed: as shown in the figure, the regulatory mechanisms are clearly revealed through comparative simulations between the control group with normal soil moisture (left) and the experimental group with 50
	Differences in surface energy exchange: In the left control group, moist soil is dominated by latent heat flux, accompanied by significant latent heat cooling, which continuously supplies water vapor to the boundary layer. In the right experimental group, dry soil is dominated by sensible heat flux, latent heat cooling is greatly weakened, and the boundary layer rapidly becomes dry and warm.
	\par
	Evolution of cold pool and convective structure: The left control group forms a strong cold pool (-1.52 K) relying on sufficient water vapor, supporting the development of a coherent squall line with concentrated and continuous precipitation bands. In the right experimental group, although the cold pool intensity is weakened (-1.21 K), the dry and warm boundary layer reduces the propagation resistance of the cold pool outflow (gust front), enhancing the lifting efficiency of the gust front. Meanwhile, the convective organization is destroyed, presenting fragmented convection, and the squall line shifts southward by approximately 280 km as a whole. Low soil moisture does not directly enhance the cooling intensity of the cold pool, but ultimately strengthens the cold pool gust front of the squall line through the pathway of "surface energy balance reconstruction → boundary layer drying and warming → optimization of dynamic characteristics of cold pool outflow".
	
	\par
	Further visual analysis of the system execution process demonstrates the complete workflow and robustness of the TianJi system in conducting end-to-end autonomous WRF squall line simulation experiments.
	\par
\begin{figure}[H]
		\centering
		\includegraphics[width=0.8\textwidth]{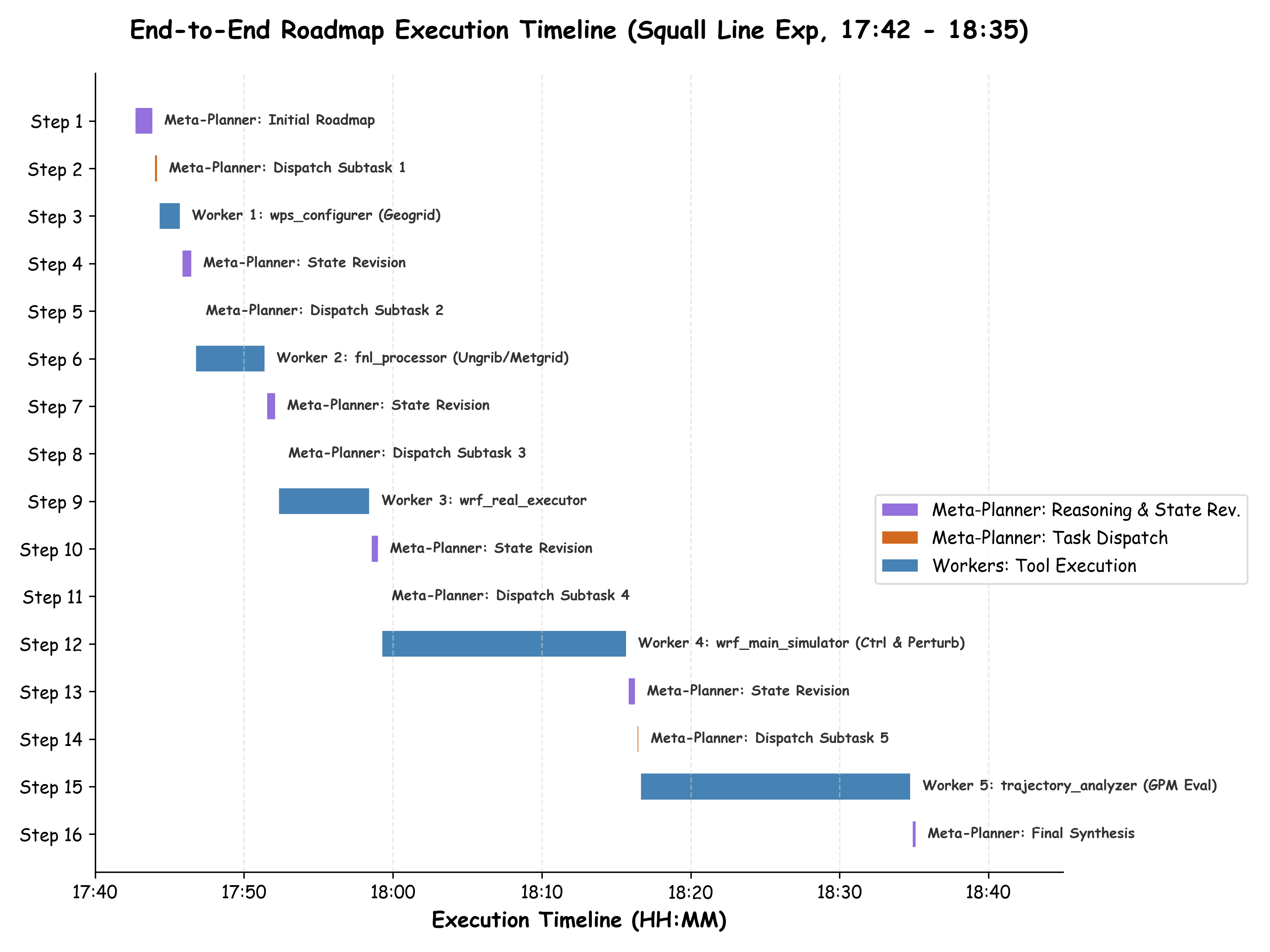} 
		\caption{End-to-End Execution Timeline for Squall Line Experiment}
		\label{img7}
	\end{figure}

	As shown in Figure\ref{img7}, within the experimental window from 17:42 to 18:35, the TianJi system, with Meta-Planner as its core, completed full-process control from initial roadmap construction to multi-Worker tool execution. The system first anchored the experimental objectives through initial roadmap planning (Step 1), and then sequentially scheduled subtasks including WPS configuration, FNL data processing, WRF initialization, main simulation, and trajectory analysis (Step 2/5/8/11/14). After each key subtask was completed, Meta-Planner automatically triggered state revision (Step 4/7/10/13/16) to verify the consistency between tool outputs and scientific logic, ensuring the experiment did not deviate from the preset hypothesis. The overall process presents the closed-loop iterative feature of "planning-scheduling-execution-verification", fully covering the entire chain of WRF simulation from preprocessing to postprocessing.

	\par
\begin{figure}[H]
		\centering
		\includegraphics[width=0.8\textwidth]{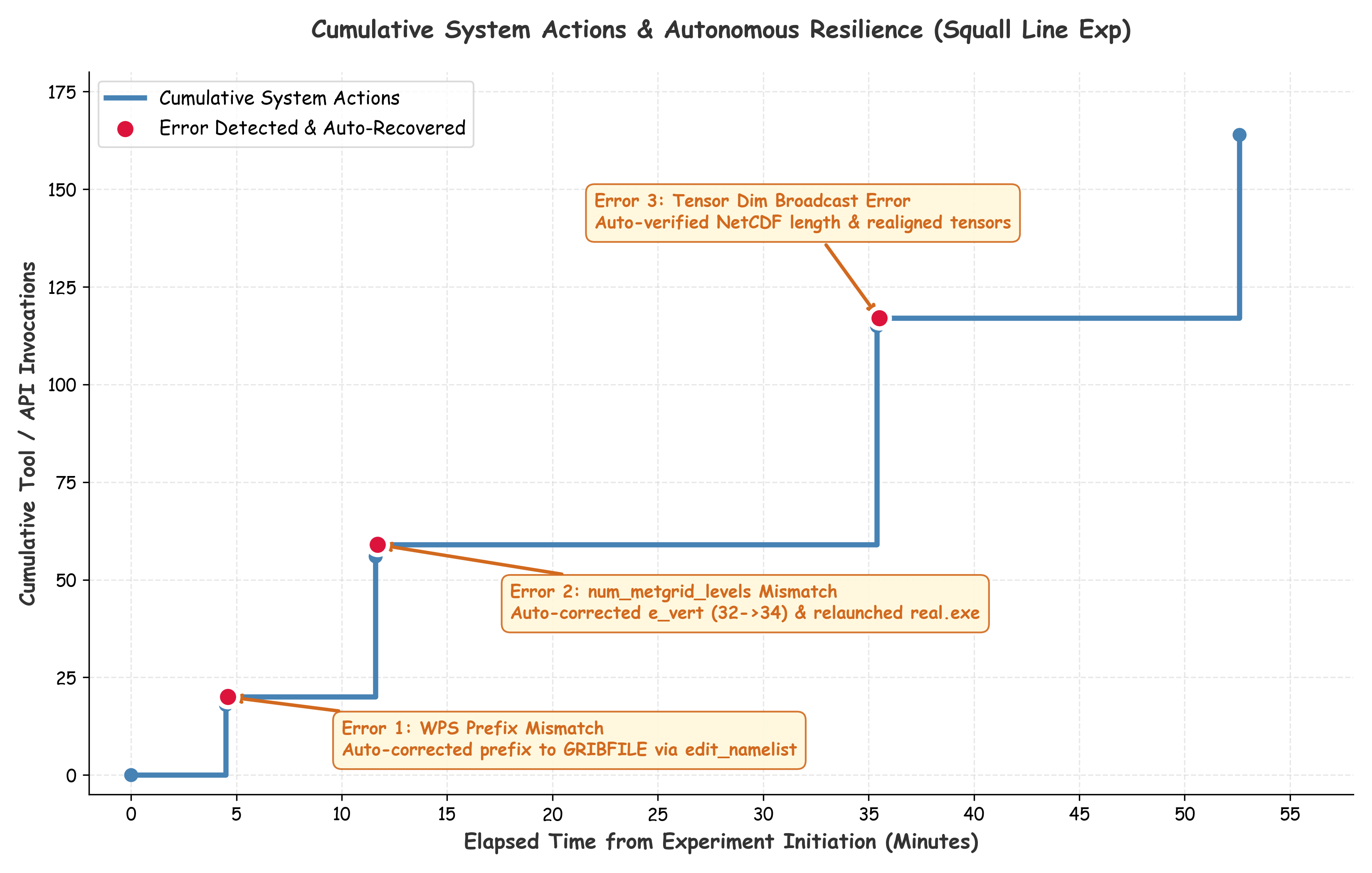} 
		\caption{Cumulative System Actions and Fault Self-Recovery}
		\label{img8}
	\end{figure}

	During execution, the system demonstrated fault self-healing capability without any human intervention. As shown in Figure\ref{img8}, within a total experimental period of approximately 55 minutes, the system triggered 164 API calls in total, and detected and automatically repaired 3 types of runtime errors during this process:
	\begin{enumerate}
		\item WPS prefix mismatch: Automatically modified the namelist to correct the prefix to GRIBFILE.
		\item metgrid level mismatch: Automatically adjusted the e\_vert parameter from 32 to 34 and reran real.exe.
		\item Tensor dimension broadcasting error: Automatically verified the NetCDF length and realigned the tensors.
	\end{enumerate}
	After each error repair, the system could resume execution seamlessly, verifying its robustness in complex meteorological simulation scenarios.
	\par
\begin{figure}[H]
		\centering
		\includegraphics[width=0.8\textwidth]{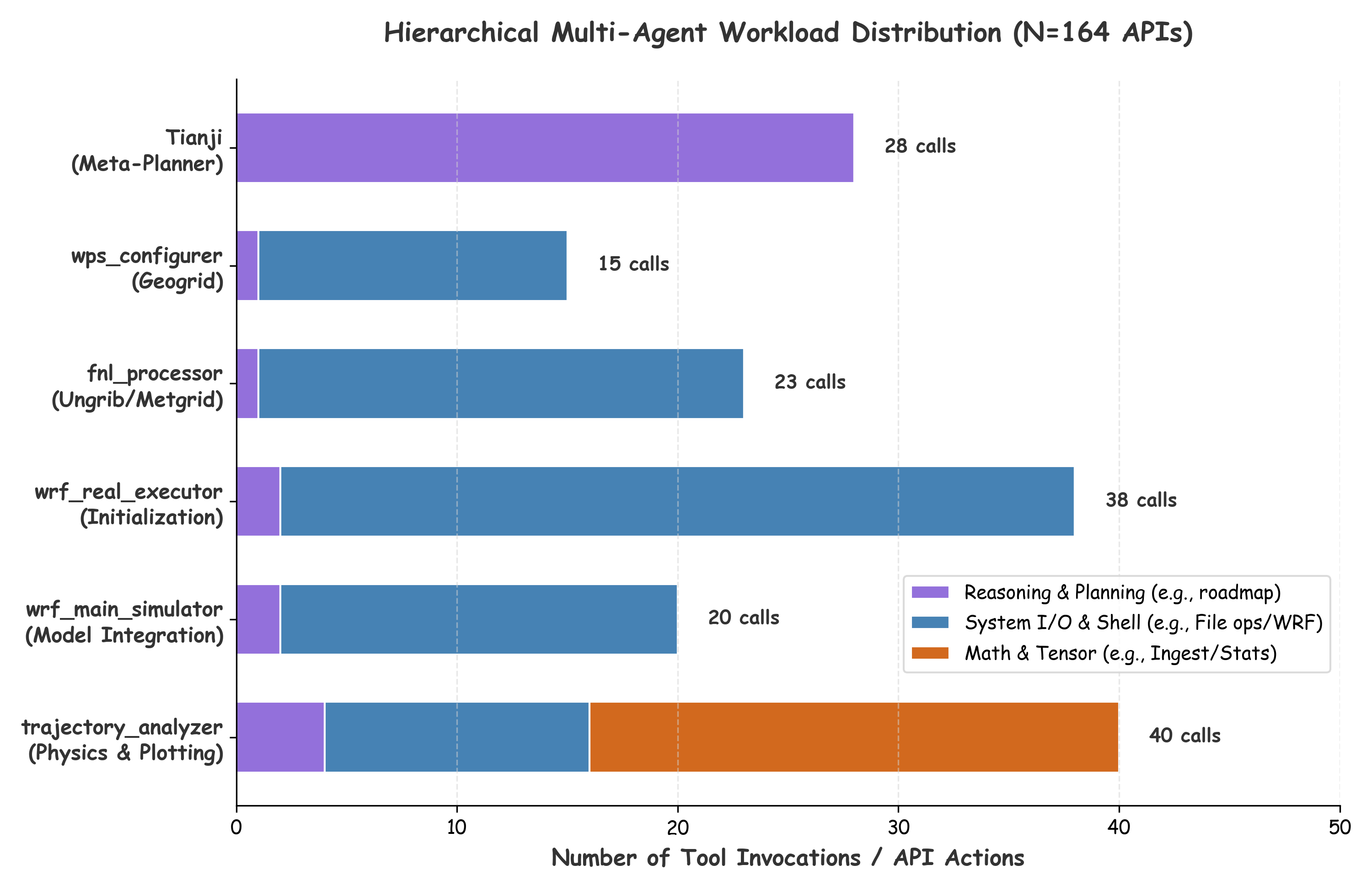} 
		\caption{Multi-Agent API Invocation Distribution}
		\label{img9}
	\end{figure}

	As shown in Figure~\ref{img9}, the system achieved clear hierarchical decoupling. Meta-Planner (TianJi) undertook 28 reasoning and planning calls, focusing on global roadmap formulation and state management without participating in underlying tool execution. Five specialized Workers shared the specific tool execution load. Among them, wrf\_real\_executor (38 calls) and trajectory\_analyzer (40 calls) bore the main system I/O and computational loads due to their involvement in model initialization and physical quantity analysis. Meanwhile, wps\_configurer (15 calls), fnl\_processor (23 calls), and wrf\_main\_simulator (20 calls) were responsible for preprocessing, model operation and other links. This modular division of labor not only ensured execution efficiency but also endowed the system with favorable scalability—no reconstruction of the core planning logic was required when adding new tools or tasks.

	\subsection{Complex Response of Typhoon Track to Sea Surface Temperature Anomalies}
	The second hypothesis validated in this study is: the impact of sea surface temperature (SST) anomalies on the track of Typhoon In-fa (2021)\cite{yang2023assessment}. The initial human hypothesis proposed an intuitive unidirectional causal chain: a global SST increase of +2°C would lead to a significant northward deflection of the typhoon track.
	\par
	The agent autonomously designed rigorous control and experimental groups, and precisely increased the SKINTEMP (sea surface skin temperature, representing SST) variable in the met\_em files containing all time steps by 2.0 K. When performing the 72-hour WRF numerical simulation, the agent not only reasonably configured core physical schemes such as Thompson microphysics, YSU boundary layer, and RRTMG radiation, but also accurately identified and disabled the SST update parameter (sst\_update = 0) to ensure that the preset temperature perturbation field remained undisturbed during the simulation—a key detail often overlooked in conventional manual modeling.
	\par
\begin{figure}[htbp]
		\centering
		\begin{subfigure}[b]{0.45\textwidth}
			\centering
			\includegraphics[width=\linewidth]{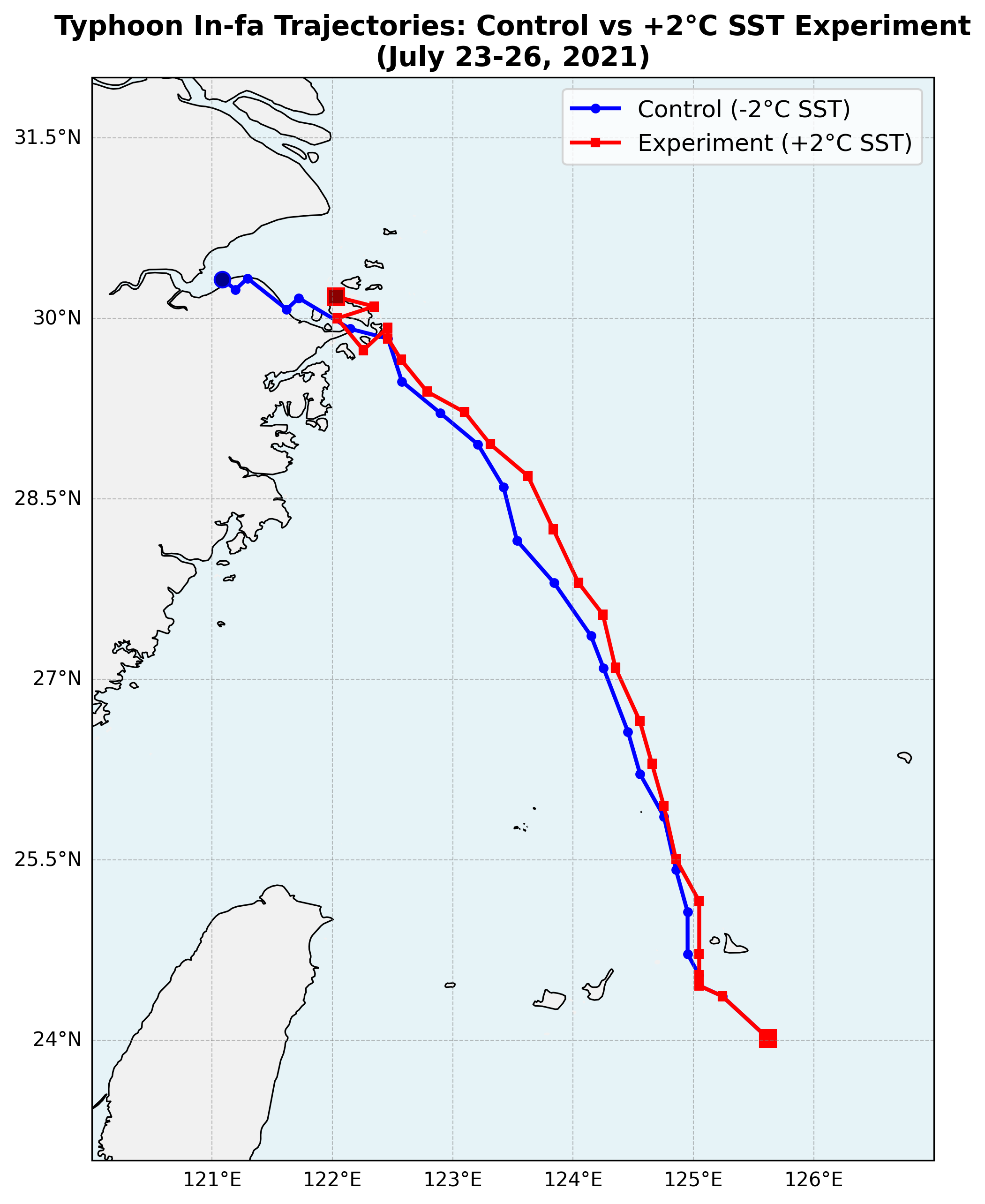}
			\caption{Comparison of typhoon tracks under SST anomalies}
			\label{img10}
		\end{subfigure}
		\hfill
		\begin{subfigure}[b]{0.45\textwidth}
			\centering
			\includegraphics[width=\linewidth]{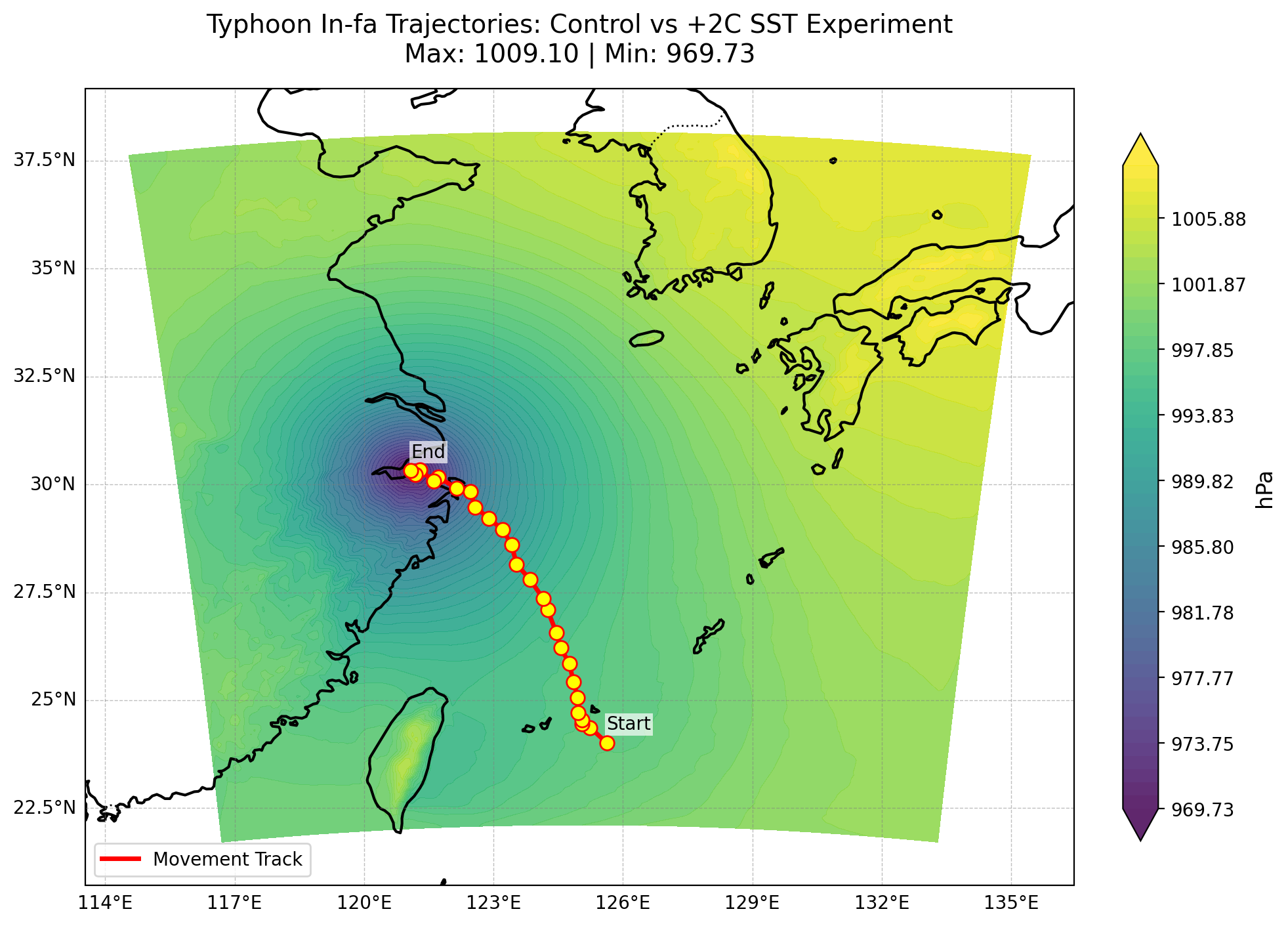}
			\caption{Evolution of central sea level pressure of Typhoon In-fa}
			\label{img11}
		\end{subfigure}
		\caption{Comparison of SST Anomaly Effects on Track and Intensity of Typhoon In-Fa}
		\label{img10_img11}
	\end{figure}

	However, the 72-hour simulation results did not confirm the expected "significant northward shift" phenomenon. The agent's autonomous diagnostic analysis fundamentally revised the causal chain set by humans, and output the results shown in Figure\ref{img10_img11}.
	
	\par
	Combined with the temporal evolution of track deviations in Figure\ref{img10}, the agent revealed that the process is not a simple linear shift: during the critical post-landing period (July 25–26), the track of the experimental group was consistently south of the control group, with a maximum southward deviation of -0.3387°.
	\par
	Combined with the typhoon intensity in Figure\ref{img11}, further physical diagnostics indicate that the +2°C SST anomaly caused a significant drop in storm central pressure (minimum sea level pressure decreased from 944.2 hPa to 924.7 hPa). The agent pointed out that the increase in SST does not directly drive the northward deflection of the typhoon track, but triggers a complex intensity-track coupling mechanism: the significantly intensified storm alters its interaction with the large-scale environmental steering flow\cite{guan2025enhanced}. By independently correcting the flawed linear hypothesis of human researchers, the agent demonstrated an unprecedented high-level physical reasoning ability among AI-driven atmospheric science tools.
	\par
\begin{figure}[H]
		\centering
		\includegraphics[width=0.8\textwidth]{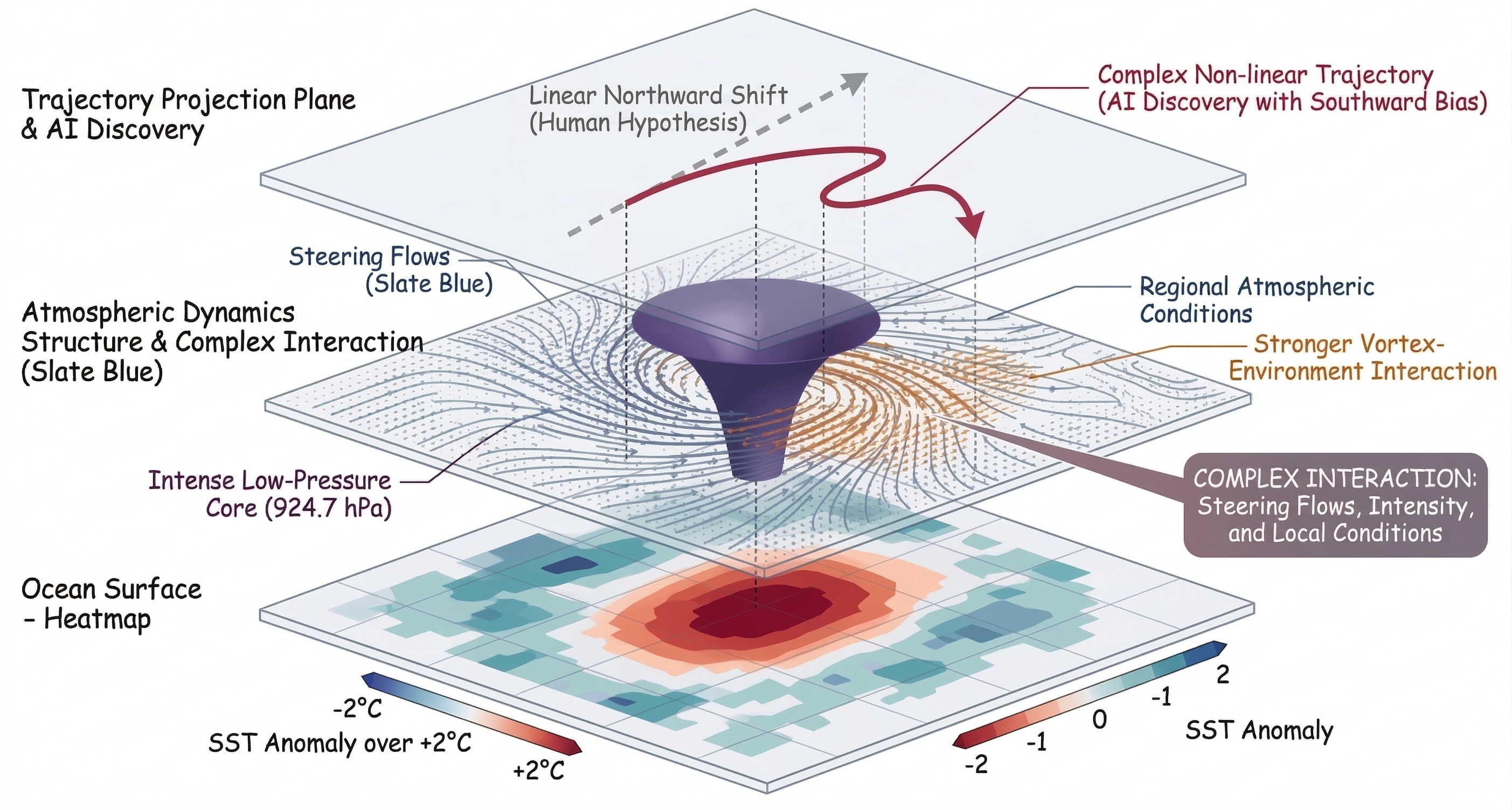} 
		\caption{Nonlinear Mechanism of SST Anomaly Regulating Typhoon Track}
		\label{img12} 
	\end{figure}

	As shown in Figure\ref{img12}, we can clearly decompose the nonlinear physical mechanism by which SST anomalies regulate typhoon tracks:
	
	\begin{enumerate}
		\item \textbf{Bottom Thermal Forcing: Energy Input from SST Anomalies} \\
		The lowermost SST anomaly heatmap shows a prominent warm SST anomaly beneath the typhoon's low-pressure center, with cold SST anomalies distributed around it. Warm SST continuously injects additional energy into the typhoon core by enhancing latent and sensible heat fluxes at the air-sea interface, strengthening the central low-pressure intensity; cold SST anomalies suppress local convective development and reshape the thermal gradient and circulation pattern around the typhoon.
		
		\item \textbf{Middle Dynamic Coupling: Complex Vortex-Environment Interactions} \\
		The atmospheric dynamic structure in the middle layer reveals the core regulatory process. The steering flow (slate blue streamlines) should dominate the linear movement of the typhoon, but the typhoon vortex intensified by warm SST undergoes strong nonlinear interactions with the environmental flow, weakening the direct traction effect of the steering flow\cite{wang2023forecasting}.
		The asymmetric thermal structure induced by local SST differences further breaks the dynamic balance, forming a multi-factor coupling effect of "steering flow + vortex intensity + local thermal conditions" (labeled \textit{COMPLEX INTERACTION}), laying the groundwork for track deviation.
		
		\item \textbf{Upper Track Response: From Linear Hypothesis to Nonlinear Discovery} \\
		In the traditional human hypothesis, the typhoon should move linearly northward along the steering flow (gray dashed line). However, actual simulations and AI autonomous findings show that the typhoon track exhibits a complex nonlinear form with a southward deviation of approximately \textbf{280 km} (red curve). The essence of this deviation is that the vortex inertia enhanced by warm SST superimposes with the southward circulation perturbation induced by cold SST, causing the typhoon to be continuously modulated by nonlinear forces during northward movement and ultimately deviate from the expected linear path. This intuitively confirms the non-monotonic, nonlinear regulatory effect of SST anomalies on typhoon tracks.
	\end{enumerate}

	\par
	To intuitively present its automated scientific research capability and engineering robustness, we further conduct a visual analysis of the execution process of this typhoon simulation experiment, revealing the operating mechanism and engineering advantages of the system from three dimensions: task timing, fault self-healing, and multi-agent load distribution.
	\par
\begin{figure}[H]
		\centering
		\includegraphics[width=0.8\textwidth]{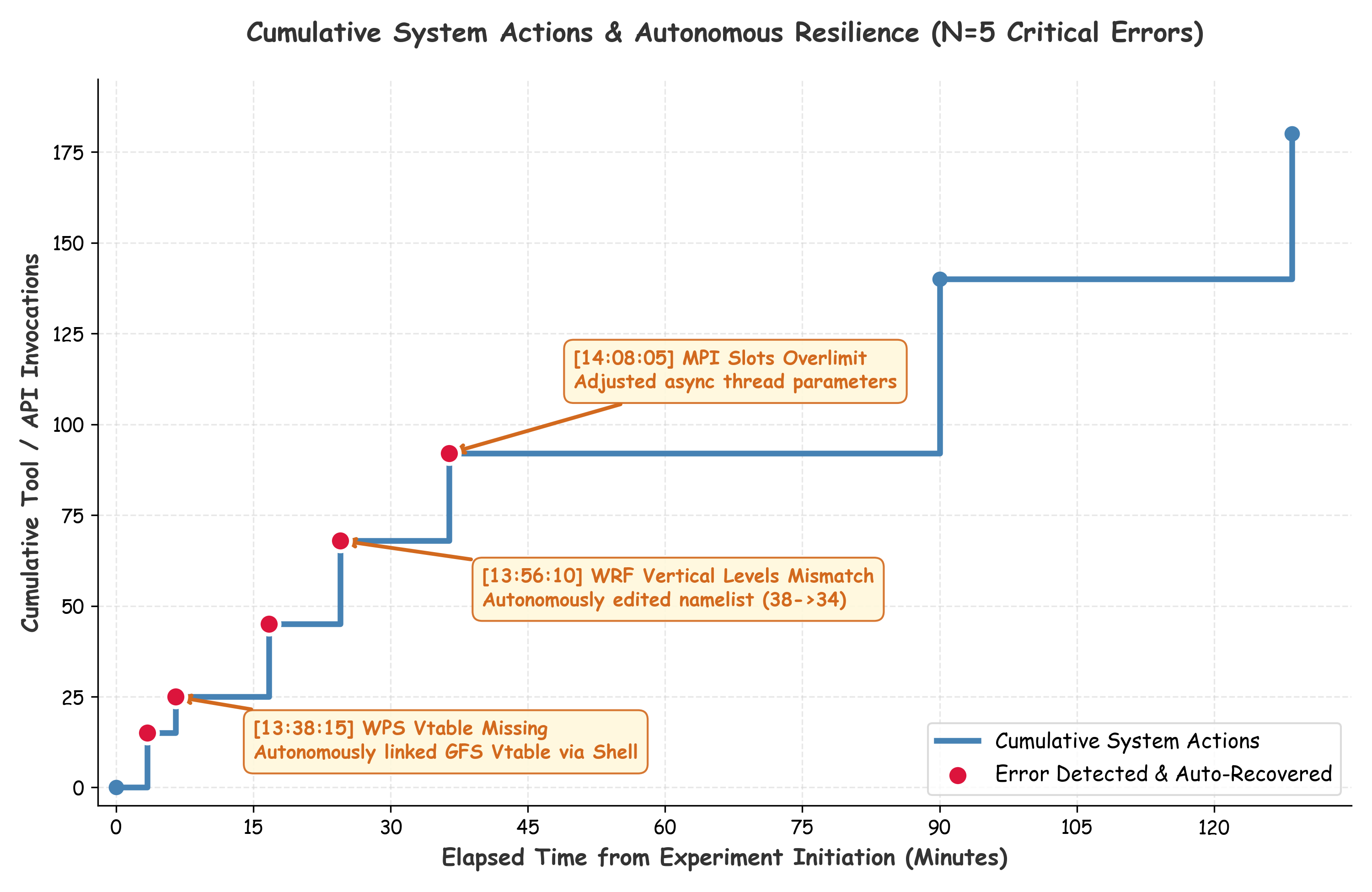} 
		\caption{Cumulative System Actions and Fault Self-Recovery}
		\label{img13} 
	\end{figure}

	As shown in Figure\ref{img13}, within an experimental period of approximately 130 minutes, the system triggered a total of 180 API calls, during which three types of critical runtime errors were detected and automatically repaired, including:
	\begin{enumerate}
		\item Missing WPS variable table (13:38:15): Automatically associated the GFS variable table via a Shell script to restore the preprocessing process.
		\item Mismatched WRF vertical levels (13:56:10): Autonomously edited the namelist to correct the number of vertical levels from 38 to 34 and reran the initialization module.
		\item MPI process overflow (14:08:05): Dynamically adjusted asynchronous thread parameters to avoid resource overflow and restarted the simulation.
	\end{enumerate}
	After each error repair, the system can resume execution seamlessly without human intervention, verifying its high robustness and autonomous regulation capability in long-duration WRF typhoon simulation scenarios.
	
	\par
\begin{figure}[H]
		\centering
		\includegraphics[width=0.8\textwidth]{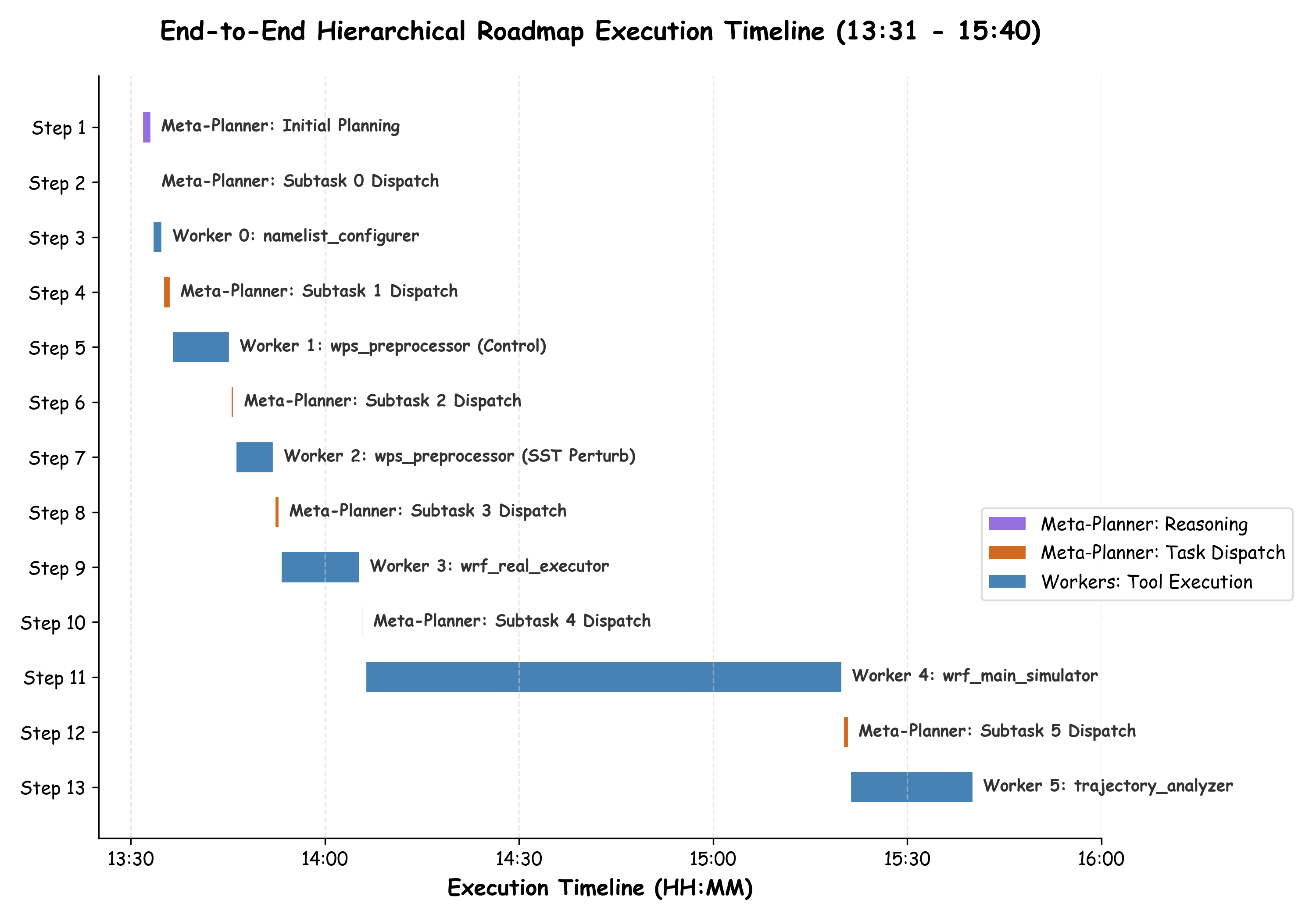} 
		\caption{End-to-End Execution Timeline for Squall Line Experiment}
		\label{img14}
	\end{figure}

	As can be seen from Figure\ref{img14}, within the experimental window from 13:31 to 15:40, the TianJi system, with Meta-Planner as its core, constructed a closed-loop workflow of planning-scheduling-execution-verification. It first anchored the scientific goal of "typhoon track response under sea surface temperature anomaly perturbation" through initial planning, and then sequentially scheduled six subtasks: namelist configuration, WPS preprocessing for the control/perturbation groups, WRF initialization, main simulation of dual experiments, and trajectory analysis. After each subtask was completed, Meta-Planner automatically triggered state verification to ensure consistency between tool outputs and the preset scientific logic and avoid deviation of the experimental direction. The overall process fully covers the entire chain of WRF typhoon simulation from configuration to post-processing, reflecting hierarchical and highly controllable end-to-end execution capability.
	\par
\begin{figure}[H]
		\centering
		\includegraphics[width=0.8\textwidth]{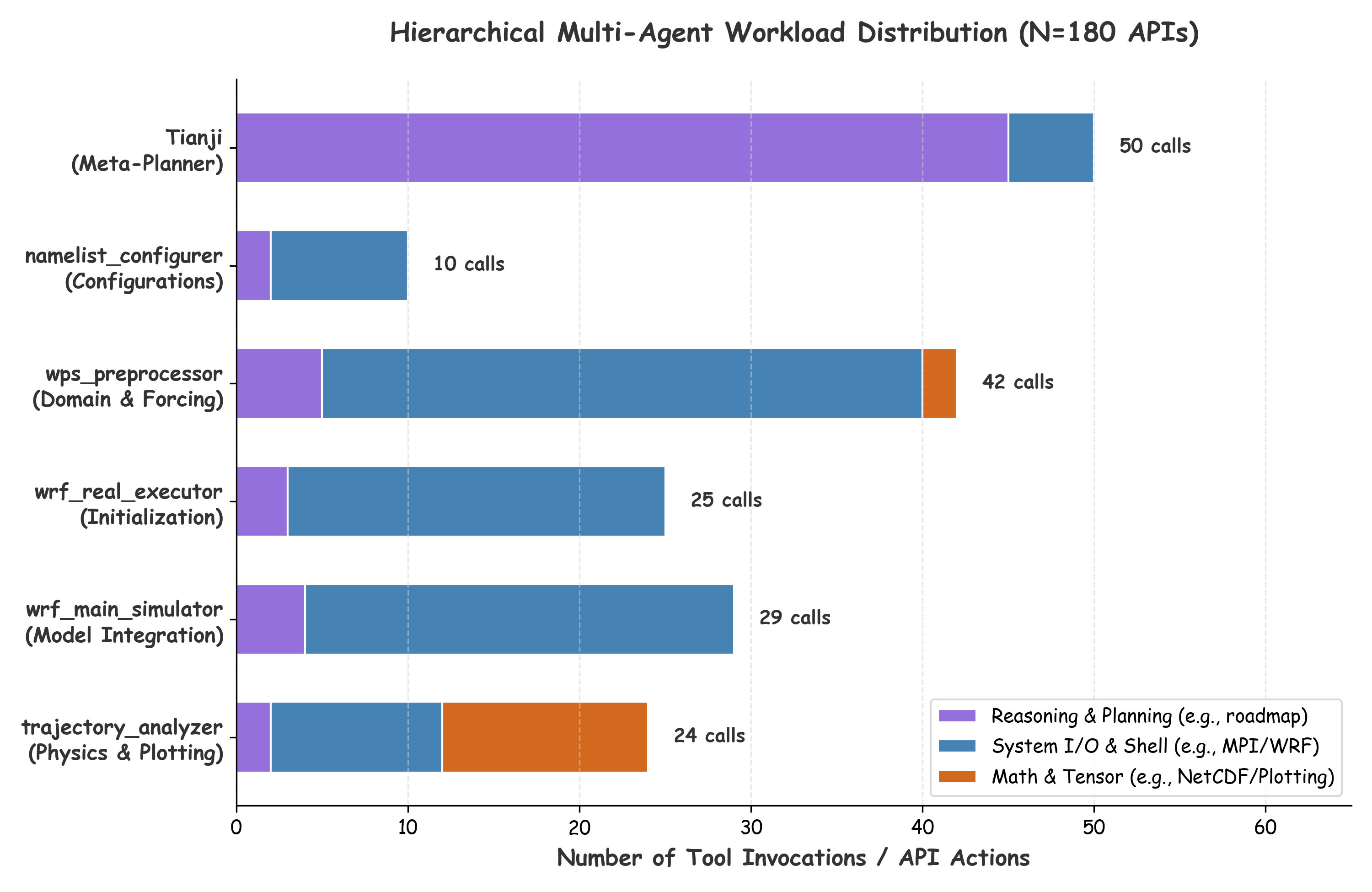} 
		\caption{Multi-Agent API Invocation Distribution}
		\label{img15}
	\end{figure}

	As shown in Figure\ref{img15}, the system achieved clear hierarchical decoupling and load balancing. Meta-Planner (TianJi) undertook 50 API calls, focusing on global reasoning and planning without participating in the execution of underlying tools. Specialized Workers shared specific execution loads: wps\_preprocessor (42 calls) and wrf\_main\_simulator (29 calls) bore the main system I/O and MPI parallel computing loads, while trajectory\_analyzer focused on tensor computing for physical quantity analysis and visualization. In terms of call types, system I/O and Shell operations accounted for the highest proportion, matching the file operation and parallel computing characteristics of WRF simulations. Meanwhile, reasoning and planning calls ensured the accurate implementation of scientific goals. This modular division of labor not only improved execution efficiency but also endowed the system with good scalability—no reconstruction of the core planning logic was required when adding new perturbation experiments or analysis modules.

	\subsection{Simple Analysis Tasks}
	This part verifies whether the system can accurately understand the requirements, successfully switch to the simple mode, and fulfill simple tasks as an assistant. Four typical meteorological diagnostic and visualization tasks were tested, and the results are summarized in Table\ref{tab:simple_tasks}.
	
	\begin{table}[htbp]
		\centering
		\caption{Summary of Simple Analysis Task Execution Results}
		\label{tab:simple_tasks}
		\begin{tabular}{m{0.35\textwidth} m{0.6\textwidth}}
			\toprule
			\textbf{Task Name} & \textbf{Execution Result (Summary)} \\
			\midrule
			Extract the minimum central pressure every 6 hours, plot the time-pressure evolution line chart, and mark the peak intensity time and value. & Successfully extracted data and generated the evolution curve (typhoon\_intensity\_evolution.png). The analysis shows that the typhoon reached peak intensity at 00:00 UTC on September 14, 2018, with a minimum central pressure of 922.769 hPa. \\
			\midrule
			Track the typhoon movement path and plot the trajectory on a high-resolution map overlaid with the sea level pressure field at peak intensity (12:00 on September 15). & Successfully located the typhoon center at each time step and extracted the sea level pressure field at 12:00 on September 15. Generated a combined map of typhoon track and pressure field with high-precision coastlines and latitude–longitude grids (typhoon\_track\_with\_slp.png). \\
			\midrule
			Calculate the total accumulated precipitation field (convective + non-convective), plot the spatial distribution using a meteorological color bar, and mark the maximum precipitation center and value with a red star. & Successfully computed the total accumulated precipitation field and generated the spatial distribution map (typhoon\_total\_precipitation.png). Precisely identified the maximum precipitation center with a total accumulated precipitation of 453.68 mm, highlighted by a red five-pointed star. \\
			\midrule
			Calculate the divergence field based on the 10-m wind field, plot the 2D filled contour map using a divergent color bar, and extract and highlight the extreme strong convergence zone. & Successfully computed the horizontal divergence field from U10 and V10 wind components and generated the filled contour map with a divergent color bar (typhoon\_divergence\_zone.png). Accurately identified the extreme strong convergence region (peak value: -0.00287 s⁻¹), highlighted by a prominent red rectangular box. \\
			\bottomrule
		\end{tabular}
	\end{table}
	
\begin{figure}[htbp]
		\centering
		\begin{subfigure}{0.48\textwidth}
			\centering
			\begin{minipage}[c]{0.7\textwidth}
				\includegraphics[width=\linewidth]{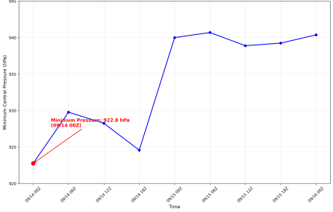}
			\end{minipage}
			\hfill
			\begin{minipage}[c]{0.25\textwidth}
				\centering \scriptsize 
				\texttt{enter\_easy\_task\_mode}\par
				\vspace{-0.7em}$\downarrow$\par \vspace{-0.7em}
				\texttt{Youtube}\par
				\vspace{-0.7em}$\downarrow$\par \vspace{-0.7em}
				\texttt{ingest\_tensor}\par
				\vspace{-0.7em}$\downarrow$\par \vspace{-0.7em}
				\texttt{locate\_feature}\par
				\vspace{-0.7em}$\downarrow$\par \vspace{-0.7em}
				\texttt{plot\_cartesian\_chart}\par
				\vspace{-0.7em}$\downarrow$\par \vspace{-0.7em}
				\texttt{generate\_response}
			\end{minipage}
			\caption{Temporal Evolution Curve of Typhoon Central Pressure}
			\label{img16}
		\end{subfigure}
		\hfill
		\begin{subfigure}{0.48\textwidth}
			\centering
			\begin{minipage}[c]{0.7\textwidth}
				\includegraphics[width=\linewidth]{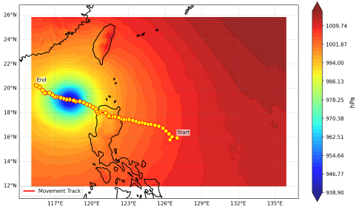}
			\end{minipage}
			\hfill
			\begin{minipage}[c]{0.25\textwidth}
				\centering \scriptsize
				\texttt{enter\_easy\_task\_mode}\par
				\vspace{-0.7em}$\downarrow$\par \vspace{-0.7em}
				\texttt{Youtube}\par
				\vspace{-0.7em}$\downarrow$\par \vspace{-0.7em}
				\texttt{ingest\_tensor}\par
				\vspace{-0.7em}$\downarrow$\par \vspace{-0.7em}
				\texttt{locate\_feature}\par
				\vspace{-0.7em}$\downarrow$\par \vspace{-0.7em}
				\texttt{plot\_spatial\_map}\par
				\vspace{-0.7em}$\downarrow$\par \vspace{-0.7em}
				\texttt{generate\_response}
			\end{minipage}
			\caption{Overlay Map of Typhoon Track and Sea Level Pressure Field}
			\label{img17}
		\end{subfigure}
		
		\vspace{0.8em}
		
		\begin{subfigure}{0.48\textwidth}
			\centering
			\begin{minipage}[c]{0.7\textwidth}
				\includegraphics[width=\linewidth]{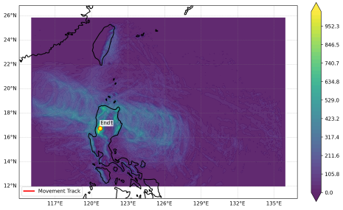}
			\end{minipage}
			\hfill
			\begin{minipage}[c]{0.25\textwidth}
				\centering \scriptsize
				\texttt{enter\_easy\_task\_mode}\par
				\vspace{-0.7em}$\downarrow$\par \vspace{-0.7em}
				\texttt{list\_directory}\par
				\vspace{-0.7em}$\downarrow$\par \vspace{-0.7em}
				\texttt{Youtube}\par
				\vspace{-0.7em}$\downarrow$\par \vspace{-0.7em}
				\texttt{ingest\_tensor}\par
				\vspace{-0.7em}$\downarrow$\par \vspace{-0.7em}
				\texttt{transform\_tensor}\par
				\vspace{-0.7em}$\downarrow$\par \vspace{-0.7em}
				\texttt{locate\_feature}\par
				\vspace{-0.7em}$\downarrow$\par \vspace{-0.7em}
				\texttt{plot\_spatial\_map}\par
				\vspace{-0.7em}$\downarrow$\par \vspace{-0.7em}
				\texttt{generate\_response}
			\end{minipage}
			\caption{Spatial Distribution Map of Total Typhoon Accumulated Precipitation}
			\label{img18}
		\end{subfigure}
		\hfill
		\begin{subfigure}{0.48\textwidth}
			\centering
			\begin{minipage}[c]{0.7\textwidth}
				\includegraphics[width=\linewidth]{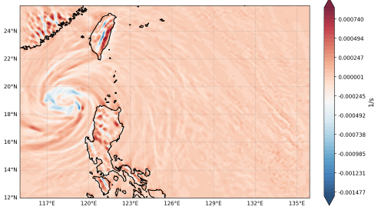}
			\end{minipage}
			\hfill
			\begin{minipage}[c]{0.25\textwidth}
				\centering \scriptsize
				\texttt{enter\_easy\_task\_mode}\par
				\vspace{-0.7em}$\downarrow$\par \vspace{-0.7em}
				\texttt{ingest\_tensor}\par
				\vspace{-0.7em}$\downarrow$\par \vspace{-0.7em}
				\texttt{transform\_tensor}\par
				\vspace{-0.7em}$\downarrow$\par \vspace{-0.7em}
				\texttt{locate\_feature}\par
				\vspace{-0.7em}$\downarrow$\par \vspace{-0.7em}
				\texttt{filter\_by\_geometry}\par
				\vspace{-0.7em}$\downarrow$\par \vspace{-0.7em}
				\texttt{plot\_spatial\_map}\par
				\vspace{-0.7em}$\downarrow$\par \vspace{-0.7em}
				\texttt{generate\_response}
			\end{minipage}
			\caption{Spatial Distribution Map of Divergence Field for 10-m Wind Field of Typhoon}
			\label{img19}
		\end{subfigure}
		
		\caption{Visualization Results of Simple Analysis Tasks}
		\label{img16_img19}
	\end{figure}

	In the diagnostic test of extreme precipitation and its dynamic triggering mechanism, the agent did not adopt the method of extracting a single precipitation variable. Based on meteorological physical principles, it autonomously synthesized convective precipitation (RAINC) and large-scale grid precipitation (RAINNC) physically, and accurately identified the intensity center of the extreme rainstorm (Figure \ref{img18}). To clarify the formation mechanism of heavy precipitation, the agent invoked data of the 10-m horizontal wind field (U and V components), combined with the model grid resolution of 9000 m, quantitatively calculated the near-surface horizontal divergence, and successfully identified a strong convergence zone with an intensity of -0.0025 s⁻¹ (Figure \ref{img19}), revealing the low-level dynamic triggering mechanism for the formation of typhoon rainstorms. The tool call sequences on the right side of Figure \ref{img18} and Figure \ref{img19} intuitively present the system execution process: after initializing the easy task mode via enter\_easy\_task\_mode, it sequentially completed directory traversal, tensor reading, physical quantity transformation, feature localization and spatial visualization, realizing the full-process automation from data to physical mechanism interpretation.
	\par
	In the task of analyzing the vertical warm-core structure of typhoons, the agent adopted the global minimum addressing algorithm to accurately locate the extreme value of sea level pressure at the typhoon center (938.9 hPa, 19.09°N, 118.17°E). Taking the typhoon center as the benchmark, the agent extracted spatial profile data of the three-dimensional height field (z) and temperature field (tc), completed the azimuthal average temperature calculation within a range of 300 km, and autonomously generated a high-resolution height-radius radial temperature contour map. This process not only achieved accurate tensor spatial slicing operations, but also reflected the agent's accurate characterization of the core meteorological physical concept of the warm-core structure of tropical cyclones\cite{lian2024impact, qi2024investigating}.
	\par
	In the task of diagnosing the full-life-cycle track of typhoons, the agent traversed the full-life-cycle simulation data time step by time step, completed the visualization of typhoon tracks overlaid with geographic coastlines (Figure \ref{img17}), extracted the minimum central pressure of the typhoon at 6-hour intervals, and generated a time-central pressure evolution line chart (Figure \ref{img16}), accurately marking the minimum pressure value at the peak intensity of the typhoon (922.8 hPa). Meanwhile, in strict accordance with meteorological operational forecasting specifications, it automatically generated a structured typhoon dynamic evolution report at 6-hour intervals, accurately recording key physical parameters. Among them, the typhoon center at the initial time was located at 15.9560°N, 126.9728°E, with a minimum central pressure of 922.7690 hPa. The workflow sequences on the right side of Figure \ref{img16} and Figure \ref{img17} show that the system entered the task mode via enter\_easy\_task\_mode, invoked tools such as ingest\_tensor, locate\_feature, plot\_cartesian\_chart and plot\_spatial\_map, and efficiently completed feature extraction and visualization output without redundant operations or human intervention.
	\par
	The above multi-dimensional diagnostic test results show that the system has gone beyond the scope of traditional programmed execution. It can autonomously interpret the analytical requirements of meteorological physical mechanisms, and reasonably decompose them into high-order tensor operations such as flow field derivation, extreme value localization, physical quantity synthesis and spatiotemporal profile extraction, realizing the full-process autonomous execution from basic processing of numerical model data to in-depth analysis of meteorological physical mechanisms.
	
\section{Conclusion}
This paper proposes TianJi, the first AI meteorologist system that can autonomously drive complex numerical models to verify physical mechanisms. The system innovatively decouples scientific research into cognitive planning and engineering execution, and constructs a two-layer multi-agent architecture consisting of hypothesis generation and hypothesis verification. In the hypothesis generation stage, the structured debate mechanism with multiple roles (host, researcher, chief scientist) effectively suppresses hallucinations and cognitive biases of large models, ensuring the physical rationality and innovation of the output hypotheses. In the verification stage, the master agent coordinates and schedules the underlying sub-agent pool through hierarchical planning, seamlessly connects with the WRF-based physical simulator, and successfully bridges the engineering gap between natural language instructions and complex Fortran numerical experiments.
\par
In practical mesoscale atmospheric dynamic scenarios, TianJi not only achieves end-to-end automated execution, but also demonstrates high-level reasoning ability to mine deep physical causal mechanisms from high-dimensional data. For example, when exploring the impact of low soil moisture on squall lines, the system not only verified the hypothesis of cold pool weakening, but also independently discovered the hidden secondary dynamic feedback mechanism of southward displacement of the main convective activity. In the verification of the impact of sea surface temperature anomalies on the track of Typhoon In-fa, it independently corrected the unidirectional linear northward shift hypothesis set by humans, and revealed the complex intensity-track coupling mechanism triggered by the drop in storm central pressure. These achievements prove that the role of AI in geoscience is successfully transforming from a purely data-driven black-box predictor to an interpretable scientific collaborator\cite{behnoudfar2026bridging}.
\par
At present, the verification scenarios of the system are mainly focused on short-term mesoscale dynamic processes, and its applicability in long-term climate simulation and multi-sphere coupling processes still needs further evaluation. Future research will focus on introducing a real-time assimilation module for multi-source observation data and expanding compatible interfaces for more Earth system models to enhance the generalization ability of the system in complex real-world environments. Through the deep integration of cutting-edge reasoning algorithms, this framework is expected to evolve into a fully autonomous closed-loop discovery system including hypothesis generation-experiment verification-hypothesis iteration, providing a feasible new paradigm for original mechanism exploration in Earth system science.

\bibliographystyle{unsrt}
\bibliography{references/reference}
	
\end{document}